# Comparative assessment of fairness definitions and bias mitigation strategies in machine learning-based diagnosis of Alzheimer's disease from MR images

Maria Eleftheria Vlontzou[1], Maria Athanasiou[1], Christos Davatzikos[2,3], Konstantina S. Nikita[1]

*Abstract*— The present study performs a comprehensive fairness analysis of machine learning (ML) models for the diagnosis of Mild Cognitive Impairment (MCI) and Alzheimer's disease (AD) from MRI-derived neuroimaging features. Biases associated with age, race, and gender in a multi-cohort dataset, as well as the influence of proxy features encoding these sensitive attributes, are investigated. The reliability of various fairness definitions and metrics in the identification of such biases is also assessed. Based on the most appropriate fairness measures, a comparative analysis of widely used pre-processing, in-processing, and post-processing bias mitigation strategies is performed. Moreover, a novel composite measure is introduced to quantify the trade-off between fairness and performance by considering the F1-score and the equalized odds ratio, making it appropriate for medical diagnostic applications. The obtained results reveal the existence of biases related to age and race, while no significant gender bias is observed. The deployed mitigation strategies yield varying improvements in terms of fairness across the different sensitive attributes and studied subproblems. For race and gender, Reject Option Classification improves equalized odds by 46% and 57%, respectively, and achieves harmonic mean scores of 0.75 and 0.80 in the MCI versus AD subproblem, whereas for age, in the same subproblem, adversarial debiasing yields the highest equalized odds improvement of 40% with a harmonic mean score of 0.69. Insights are provided into how variations in AD neuropathology and risk factors, associated with demographic characteristics, influence model fairness.

## I. INTRODUCTION

In recent years, the study of Alzheimer's Disease (AD) has attracted significant attention due to the disease's association with the onset of cognitive and motor impairments that profoundly affect quality of life and are among the leading causes of morbidity and disability in the elderly population [1]. It is estimated that 32 million people globally are affected by AD-related dementia, with this number rising to 416 million when considering earlier disease stages, such as Mild Cognitive Impairment (MCI) [2]. Therefore, advancing research in this field is becoming an urgent priority, requiring a shift in focus from merely diagnosing diseases to unravelling the complexity of their heterogenous neuroanatomical and pathophysiological variations [3], which may not only arise from diverse anatomies, clinical phenotypes or genetic traits, but also from demographic factors, including associations with age, gender or race, as well as from the insufficient representation of certain subgroups in the data [4].

However, the variations in disease neuropathology, progression rates, and risk factors stemming from demographic characteristics have been insufficiently studied [5]. Several studies have reviewed existing evidence on gender and racial disparities in Alzheimer's Disease (AD) and non-AD dementias, highlighting the limited research so far in terms of geographical- and gender-related differences [6]. For instance, evidence suggests that women are more likely to develop AD and show faster aging-related cognitive deterioration than men [5]. At the same time, there are contradicting findings related to the effect of established risk factors, such as the APOE4 gene, across races. Studies have found that APOE4 is not a significant risk factor for AD among elderly African Americans, as compared to its role in white populations [1]. In the context of machine learning (ML) based studies, these disparities can introduce biases that lead to unfair outcomes, such as the under-diagnosis or over-diagnosis of specific demographic populations. Thus, unravelling such variations and prioritizing diagnostic fairness and trustworthiness in artificial intelligence (AI) clinical decision support systems, is vital for their adoption in clinical practice, as well as for promoting the equitable allocation of medical resources.

Several studies have explored the unique challenges and requirements of fairness-aware ML approaches for detecting and mitigating biases in medical contexts [5], [7], highlighting the importance of a comprehensive approach for defining fairness metrics, identifying all potential sources of bias, and addressing the trade-off between fairness and predictive performance. Various definitions have been proposed for fairness in ML, each capturing different dimensions of bias. Group fairness evaluates whether a model performs equally across demographic subgroups [5], while individual fairness requires that similar individuals with respect to non-sensitive attributes receive similar treatment by the model [8]. Counterfactual fairness captures whether model performance is maintained in case only the sensitive attribute is altered [9]. Min-max fairness addresses the trade-off between fairness and performance by minimizing the error rate of the least privileged subgroup, instead of trying to achieve equal error rates across subgroups. Apart from exploring different fairness definitions, mitigating bias necessitates understanding its sources, which may arise due to imbalances in the data, human-introduced labelling biases, or the existence of proxy features that may indirectly reveal information about sensitive attributes [10].

[1]ME. Vlontzou, M. Athanasiou, K. Nikita are with Faculty of Electrical and Computer Engineering, National Technical University of Athens, Greece  mvlontzou@biosim.ntua.gr, mathanasiou@biosim.ntua.gr, knikita@ece.ntua.gr

[2]C. Davatzikos is with Center for Biomedical Image Computing and Analytics and with [3] Department of Radiology, University of Pennsylvania, Philadelphia, PA, USA christos.davatzikos@pennmedicine.upenn.edu

Assessing fairness in medical contexts requires the appropriate selection of fairness definitions and corresponding metrics, depending on whether the goal is equal allocation of resources or equitable model performance across subgroups' tasks [11]. For instance, demographic parity, which ensures equality of decision rates among subgroups without considering the ground truth label, may be appropriate for resource distribution (e.g., organ transplant allocation) but unsuitable for diagnostic tasks (e.g., distinguishing among healthy individuals and MCI or AD), where Equalized Odds, ensuring parity in True Positive Rates (TPR) and False Positive Rates (FPR) between subgroups, would be more relevant. Moreover, given that a crucial aspect of fairness assessment relates to the significant trade-off between fairness and predictive performance, it is essential to evaluate models using metrics that jointly assess fairness and utility [7].

In the context of MCI and AD diagnosis, several studies have examined fairness across subgroups defined by gender, age, or race, primary focusing on race and gender-related biases [12], [13]. Some have extended this analysis to age-related biases, either by examining fairness in terms of gender, race, age, and their intersections, or by exploring bias mitigation strategies through data preprocessing and hyperparameter tuning [4], [14]. However, the trade-off between fairness and predictive performance has remained largely unquantified, limiting direct comparisons across models and mitigation techniques. While some studies have emphasized the inherent conflict between fairness and performance [7], others have argued that appropriate data preprocessing can reconcile these objectives [15]. Efforts to quantify this trade-off include the introduction of accurate fairness to measure the balance between individual fairness and accuracy criteria [16], and the use of Minimax Pareto frameworks for optimizing both individual and group fairness [17].

The present study aims to bridge this gap by performing a systematic fairness analysis to investigate the fairness-performance trade-off in ML models for MCI and AD diagnosis, focusing on biases introduced by age, gender, or race. The main contributions of this work are the following:

- It performs a large-scale fairness analysis on neuroimaging-derived data of Cognitively Normal (CN) participants and MCI, AD patients from multiple studies, with the aim of enhancing the robustness of the obtained findings. The analysis considers age, gender, and race, as well as proxy features identified through explainability methods.
- Moreover, it comparatively assesses the effectiveness of pre-processing, in-processing, and post-processing bias mitigation techniques for each sensitive attribute using both group and counterfactual fairness metrics to determine the most suitable fairness definition for medical diagnostic applications.
- It introduces a composite metric for quantifying the trade-off between fairness and performance, facilitating model comparison and optimization of both performance and fairness in the context of medical diagnostics.

## II. METHODS

### A. Data

Data from the iSTAGING consortium [18], comprising volumetric ROIs from T1-weighted MRI brain scans of 9155 participants, were utilized for the model's development and the implementation of the present fairness analysis. Data included cross-sectional baseline scans of 6829 cognitively healthy individuals, 1191 MCI patients, and 1135 AD patients, which were corrected for intensity inhomogeneities, segmented using the MUSE multi-atlas segmentation method [19] and then harmonized using the multi-variate ComBAT method along with generalized additive models [20] to remove the effect of site, while age and gender-related differences were preserved.

The dataset was relatively balanced in terms of gender, consisting of 54.8% females and 45.2% males, with males accounting for 42.8% of healthy individuals, 56.6% of MCI patients, and 47.8% of AD patients. A notable imbalance was observed in race, with 87.7% of participants identified as white and only 12.3% as black. To facilitate a focused fairness analysis, participants with unknown race or those from underrepresented groups, such as Asians, were excluded, limiting the analysis to black and white individuals. White participants constituted 86.4% of CN, 94.3% of MCI, and 89% of AD patients. Participants' age ranged from 49 to 103 years, with a median of 69 years, which was used as the threshold for defining two age groups in the fairness analysis. Individuals under 69 years represented 60.5% of CN, 27.5% of MCI, and 15.9% of AD patients. The sensitive attributes of gender, age, and race were excluded from the models' input space during classification, except for the cases of adversarial debiasing and counterfactual fairness calculation, in which the respective sensitive attribute was included.

### B. Overview of the proposed fairness analysis approach

The conceptual framework of the proposed fairness analysis approach is depicted in Fig. 1. It incorporates the development of classifiers based on the adoption of an ensemble learning method for the classification of healthy, MCI, and AD participants, the application of different bias mitigation techniques, and their evaluation through various fairness and predictive performance metrics as well as through the introduction of a composite metric for assessing the trade-off between fairness and predictive performance. The above components are described in detail in the following subsections.

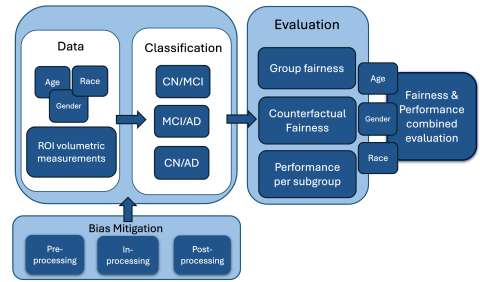

Fig. 1: Schematic overview of the classification, bias mitigation and fairness evaluation workflow

*C. Classification*

For the multiclass classification problem defined by the CN, MCI, and AD classes, the One Versus One (OVO) decomposition scheme was employed, combined with an ensemble learning method, to address the effects of class imbalance. The majority class was partitioned into two training subsets, each of which was further decomposed into three binary classification subsets. Subsequently, six Support Vector Machine (SVM) classifiers were trained, and their predictions were aggregated using a voting strategy that assigned each instance to the class with the highest predicted probability. A 5x4-fold nested cross-validation scheme was applied for hyperparameters' optimization, with selection criteria based on the weighted F1-score and balanced accuracy [21].

*D. Fairness evaluation*

Fairness was evaluated for each sensitive attribute using both group and counterfactual fairness. Group fairness was assessed through standard metrics, including demographic parity and equalized odds ratios, as well as accuracy parity and F1-score parity ratios, to capture performance disparities. Moreover, true positive (TPR), false positive (FPR), false negative (FNR) and true negative rate (TNR) ratios were calculated to provide detailed insights about the effect of each sensitive attribute on the three pairs of classes. Unlike most fairness studies [5], which rely on the absolute differences in fairness metric scores between subgroups, this study employed ratios - calculated as the minimum to maximum metric value - to provide proportional and comparative insights.

To complement group fairness evaluation, counterfactual fairness was assessed for each sensitive attribute by creating a second training set with perturbed sensitive attribute values. Counterfactual fairness was measured by computing the percentage of predictions that remained consistent under the perturbation of the respective sensitive attribute [9], calculated both at the subgroup level and across the entire dataset. Among the applied bias mitigation techniques described in the following subsection, adversarial debiasing was excluded from counterfactual fairness evaluation, since the sensitive attributes could not be included in the input space, due to the adversary's task of predicting them.

In addition to fairness, performance utility metrics, including accuracy, F1-score, and TPR, FPR, FNR, TNR, were computed per subgroup and across all groups for the three pairs of classes, providing a complete overview of performance disparities. All metrics were calculated as the average across folds in the applied 5-fold cross-validation scheme.

To systematically examine the trade-off between model fairness and performance, a composite metric, tailored for medical diagnosis contexts, was introduced. Among group fairness metrics, equalized odds ratio is considered particularly suitable, as it accounts for both TPR and FPR ratios, ensuring equal performance among subgroups. F1-score, which optimizes precision and recall, serves as an appropriate measure of model performance. Thus, to address both fairness and predictive performance aspects and quantify their trade-off, the harmonic mean of equalized odds ratio and weighted F1-score was calculated. The harmonic mean formulation ensures that the metric obtains high values, only when both the F1-score and the equalized odds ratio are high, thus reflecting acceptable levels of fairness and predictive performance. If either of the two metrics is low, the harmonic mean is strongly affected, even if the other component is high. This characteristic makes the proposed composite metric a robust indicator of the trade-off between fairness and utility. It is calculated as:

$$\text{Harmonic mean} = \frac{2}{\frac{1}{\text{Equalized odds ratio}} + \frac{1}{\text{Weighted F1 score}}}, \quad (1)$$

in which equalized odds ratio is the average of TPR and FPR ratio:

$$\text{Equalized odds ratio} = \frac{\frac{\text{TPR (min)}}{\text{TPR (max)}} + \frac{\text{FPR (min)}}{\text{FPR (max)}}}{2} \quad (2)$$

*E. Bias mitigation*

To mitigate bias, three distinct strategies were deployed and comparatively assessed, representing interventions at the pre-processing, in-processing, and post-processing stage.

*1) Pre-processing:* Pre-processing methods focus on appropriately processing the data before training to remove bias. In the present study, a linear covariates adjustment approach was employed to remove the effects of age, gender, and race. For each sensitive attribute, the covariates' correlation with the ROI volumes was predicted with linear regression in CN participants, and all ROI volumetric measurements were subsequently residualized. This method also allowed eliminating the effect of total brain volume, which was identified as the only proxy feature, by including it as a predictor alongside each sensitive attribute. This adjustment facilitated retaining only the disease-related neuroanatomical information of the ROI volumes in the data, while discarding associations with sensitive and proxy features.

For the identification of proxy features, which may indirectly encode age, race, or gender, the SHapley Additive exPlanations (SHAP) method was used. SHAP is an interpretability method that quantifies features' importance based on their marginal contribution to model predictions. In this study, SHAP was leveraged to rank global feature attributions in the auxiliary task of predicting each sensitive attribute [10].

*2) In-processing:* In-processing methods aim to mitigate biases by modifying the model architecture or incorporating fairness objective functions and constraints [5]. In the present study, adversarial debiasing was employed, a popular technique that trains a classifier to maximize prediction performance and at the same time minimize an adversary's ability to predict the sensitive attribute [22]. To implement this, each SVM classifier of the ensemble learning scheme was substituted by an adversarial debiasing model. The adversarial debiasing model implementation from AIF360 fairness toolbox was used and hyperparameters' tuning was performed via 5-fold cross-validation, stratified by both the class label and sensitive attribute. Hyperparameters, including the number of epochs, batch size, number of hidden units of the classifier, and the adversary loss weight, which controls the learning rate decay [23], were optimized based on the harmonic mean of the F1-score and equalized odds ratio. For each binary classification problem (i.e., CN/MCI, MCI/AD, and CN/AD) and for each sensitive attribute, the optimal hyperparameters' combination was utilized for training the adversarial debiasing model.

*3) Post-processing:* Post-processing methods involve calibrating the models' output to mitigate bias. In this study, the Reject Option-based Classification (ROC) method from Holistic AI [1] was applied, introducing a region of posterior probability estimates around the model's decision boundary, in which the predictions are uncertain. Instances falling within this region were relabelled by identifying the threshold maximizing fairness with respect to equalized odds [24]. In particular, after training an ensemble of SVM models for each binary subproblem, ROC was applied to identify the optimal threshold for relabelling uncertain instances, that was selected among 100 candidate thresholds based on equalized odds. A 5-fold cross-validation scheme was used to determine the optimal lower and upper classification thresholds, defining the optimal threshold search space for ROC.

## III. RESULTS

### A. Fairness evaluation

The ensemble SVM classifier, prior to the application of bias mitigation techniques, achieved overall discriminative performance of 91% weighted F1-score and 84% balanced accuracy. Tables I-III and Fig. 2-4 present the classifiers' fairness evaluation results for each sensitive attribute, before and after the application of the different bias mitigation strategies. The reported metrics represent the average values obtained through the 5-fold cross-validation scheme.

In terms of race, when no bias mitigation technique was applied, demographic parity ratio exhibited relatively low values, especially in the CN/MCI pair of classes, though the application of ROC significantly improved this metric, except for the case of the MCI/AD binary subproblem, as shown in Table I. Before bias mitigation, equalized odds ratio yielded low values during the distinction between MCI and AD, but obtained relatively high values in the other two subproblems, which were further improved by the applied bias mitigation techniques. These disparities can be mostly attributed to differences in the TPR, since white participants had lower TPR than black participants in the CN/MCI subproblem, while the opposite was observed in the MCI/AD subproblem. Notably, adversarial debiasing and ROC led to a significant improvement of the equalized odds ratio in the MCI/AD subproblem. Despite the obtained low fairness scores in some cases, accuracy parity and F1-score parity remained consistently high before and after the application of bias mitigation techniques. The comparative assessment of the obtained F1-core, equalized odds ratio, and their harmonic mean, depicted in Fig. 2, indicated the absence of significant bias in the CN/MCI and CN/AD subproblems, while highlighting the low values of equalized odds ratio in MCI/AD, which negatively impacted the harmonic mean metric. Across all bias mitigation techniques, similar harmonic mean values were observed for the CN/MCI and CN/AD subproblems, while adversarial debiasing and ROC substantially improved the harmonic mean in MCI/AD, demonstrating maximization of both the F1-score and equalized odds ratio.

[1] https://holisticai.readthedocs.io/en/latest/

TABLE I: Fairness metrics before and after the application of bias mitigation techniques for race sensitive attribute.

| Race | | No Mitigation | Race Correction | Race & Total Brain Volume Correction | Adversarial Debiasing | Reject Option Classification |
|---|---|---|---|---|---|---|
| **Demographic Parity** | CN/MCI | 0.43 ±0.04 | 0.47 ±0.05 | 0.46 ±0.05 | 0.41 ±0.11 | 0.73 ±0.16 |
| | MCI/AD | 0.61 ±0.13 | 0.66 ±0.14 | 0.70 ±0.13 | 0.69 ±0.07 | 0.63 ±0.10 |
| | CN/AD | 0.75 ±0.13 | 0.77 ±0.16 | 0.73 ±0.13 | 0.66 ±0.08 | 0.91 ±0.04 |
| **Equalized Odds** | CN/MCI | 0.68 ±0.11 | 0.76 ±0.11 | 0.77 ±0.10 | 0.77 ±0.18 | 0.81 ±0.08 |
| | MCI/AD | 0.43 ±0.05 | 0.43 ±0.04 | 0.43 ±0.02 | 0.62 ±0.18 | 0.63 ±0.18 |
| | CN/AD | 0.79 ±0.10 | 0.78 ±0.12 | 0.78 ±0.12 | 0.82 ±0.16 | 0.86 ±0.07 |
| **Balanced Accuracy Parity** | CN/MCI | 0.86 ±0.07 | 0.93 ±0.03 | 0.97 ±0.02 | 0.93 ±0.06 | 0.92 ±0.06 |
| | MCI/AD | 0.93 ±0.04 | 0.93 ±0.03 | 0.93 ±0.01 | 0.94 ±0.03 | 0.97 ±0.01 |
| | CN/AD | 0.97 ±0.02 | 0.95 ±0.03 | 0.98 ±0.03 | 0.98 ±0.01 | 0.98 ±0.01 |
| **F1-score Parity** | CN/MCI | 0.98 ±0.01 | 0.99 ±0.01 | 0.99 ±0.01 | 0.98 ±0.01 | 0.98 ±0.01 |
| | MCI/AD | 0.94 ±0.03 | 0.95 ±0.04 | 0.93 ±0.05 | 0.95 ±0.02 | 0.97 ±0.01 |
| | CN/AD | 0.99 ±0.01 | 0.99 ±0.01 | 0.99 ±0.01 | 0.98 ±0.01 | 0.98 ±0.01 |

The fairness analysis for gender showed absence of bias, especially in the CN/MCI and CN/AD pairs, where all fairness metrics (Table II) were significantly high. Disparities were observed only in the MCI/AD subproblem, where equalized odds received significantly lower values compared to the other subproblems across all bias mitigation strategies, except for ROC post-processing method, which greatly improved the equalized odds ratio. ROC also yielded the highest utility metrics in the case of CN/MCI and CD/AD subproblems. As shown in Fig. 3, the application of ROC maximized the weighted F1-score and achieved a similar equalized odds ratio score to other bias mitigation techniques in CN/MCI and CN/AD, while performing best in terms of equalized odds in MCI/AD, thus maximizing the harmonic mean metric, which demonstrated its ability to balance fairness and utility.

For age subgroups, demographic parity exhibited low values across all binary subproblems and only presented an improvement after the application of linear correction at the pre-processing stage, as shown in Table III. Equalized odds ratio yielded low values prior to bias mitigation, particularly for the MCI/AD subproblem, but presented an improvement after the application of the linear correction and adversarial debiasing. The obtained equalized odds results were influenced by disparities in TPR and FPR ratios, with participants over 69 years showing higher FPR, which indicated greater susceptibility to overdiagnosis, and higher TPR, suggesting better disease detection. Before the application of any bias mitigation technique, significant disparities were also observed across all subproblems in the FNR ratio, where participants under 69 years exhibited higher FNR. Similar to the other sensitive attributes, balanced accuracy parity and F1-score parity ratios were relatively high and did not reflect the existing bias. Based on the obtained harmonic mean of the F1-score and equalized odds ratio depicted in Fig. 4, adversarial debiasing was highlighted as the best performing bias mitigation technique in the CN/MCI and CN/AD subproblems, while for MCI/AD, it was shown to maximize equalized odds ratio, but with a significant performance trade-off. In the MCI/AD subproblem, linear correction based on age at the preprocessing stage yielded the highest harmonic mean, effectively striking a balance between the F1-score and equalized odds. Notably, adversarial debiasing improved overall fairness in the MCI/AD subproblem, but at the cost of increasing FNR and decreasing TPR for both age groups, thus leading to potential underdiagnosis.

TABLE II: Fairness metrics before and after the application of bias mitigation techniques for gender sensitive attribute.

| Gender | | No Mitigation | Gender Correction | Gender & Total Brain Volume Correction | Adversarial Debiasing | Reject Option Classification |
|---|---|---|---|---|---|---|
| **Demographic Parity** | CN/MCI | 0.91 ± 0.04 | 0.93 ± 0.05 | 0.91 ± 0.04 | 0.83 ± 0.04 | 0.72 ± 0.06 |
| | MCI/AD | 0.79 ± 0.04 | 0.79 ± 0.04 | 0.78 ± 0.06 | 0.80 ± 0.04 | 0.82 ± 0.02 |
| | CN/AD | 0.95 ± 0.04 | 0.92 ± 0.04 | 0.93 ± 0.04 | 0.90 ± 0.05 | 0.88 ± 0.05 |
| **Equalized Odds** | CN/MCI | 0.93 ± 0.05 | 0.91 ± 0.04 | 0.91 ± 0.03 | 0.93 ± 0.04 | 0.90 ± 0.06 |
| | MCI/AD | 0.47 ± 0.00 | 0.55 ± 0.17 | 0.47 ± 0.02 | 0.48 ± 0.02 | 0.74 ± 0.11 |
| | CN/AD | 0.94 ± 0.03 | 0.91 ± 0.04 | 0.91 ± 0.03 | 0.93 ± 0.04 | 0.89 ± 0.06 |
| **Balanced Accuracy Parity** | CN/MCI | 0.95 ± 0.03 | 0.95 ± 0.02 | 0.95 ± 0.01 | 0.98 ± 0.02 | 0.99 ± 0.01 |
| | MCI/AD | 0.98 ± 0.02 | 0.97 ± 0.02 | 0.96 ± 0.02 | 0.98 ± 0.02 | 0.97 ± 0.02 |
| | CN/AD | 0.97 ± 0.01 | 0.97 ± 0.02 | 0.97 ± 0.01 | 0.97 ± 0.03 | 0.97 ± 0.06 |
| **F1-score Parity** | CN/MCI | 0.97 ± 0.02 | 0.96 ± 0.01 | 0.97 ± 0.02 | 0.97 ± 0.02 | 0.99 ± 0.01 |
| | MCI/AD | 0.98 ± 0.02 | 0.98 ± 0.02 | 0.97 ± 0.02 | 0.98 ± 0.02 | 0.98 ± 0.02 |
| | CN/AD | 0.98 ± 0.02 | 0.97 ± 0.01 | 0.98 ± 0.02 | 0.96 ± 0.02 | 0.98 ± 0.01 |

TABLE III: Fairness metrics before and after the application of bias mitigation techniques for age sensitive attribute.

| Age | | No Mitigation | Age Correction | Age & Total Brain Volume Correction | Adversarial Debiasing | Reject Option Classification |
|---|---|---|---|---|---|---|
| **Demographic Parity** | CN/MCI | 0.42 ± 0.04 | 0.55 ± 0.04 | 0.53 ± 0.03 | 0.42 ± 0.04 | 0.47 ± 0.25 |
| | MCI/AD | 0.59 ± 0.03 | 0.68 ± 0.05 | 0.70 ± 0.04 | 0.56 ± 0.12 | 0.50 ± 0.11 |
| | CN/AD | 0.38 ± 0.04 | 0.52 ± 0.04 | 0.49 ± 0.03 | 0.31 ± 0.08 | 0.35 ± 0.31 |
| **Equalized Odds** | CN/MCI | 0.63 ± 0.04 | 0.74 ± 0.02 | 0.74 ± 0.04 | 0.78 ± 0.17 | 0.65 ± 0.21 |
| | MCI/AD | 0.42 ± 0.02 | 0.56 ± 0.15 | 0.52 ± 0.06 | 0.59 ± 0.10 | 0.53 ± 0.16 |
| | CN/AD | 0.67 ± 0.04 | 0.79 ± 0.02 | 0.78 ± 0.02 | 0.73 ± 0.20 | 0.63 ± 0.23 |
| **Balanced Accuracy Parity** | CN/MCI | 0.94 ± 0.04 | 0.98 ± 0.01 | 0.97 ± 0.02 | 0.97 ± 0.02 | 0.98 ± 0.01 |
| | MCI/AD | 0.93 ± 0.02 | 0.97 ± 0.03 | 0.97 ± 0.03 | 0.94 ± 0.03 | 0.91 ± 0.05 |
| | CN/AD | 0.90 ± 0.01 | 0.91 ± 0.02 | 0.92 ± 0.03 | 0.93 ± 0.03 | 0.96 ± 0.01 |
| **F1-score Parity** | CN/MCI | 0.80 ± 0.01 | 0.87 ± 0.01 | 0.89 ± 0.01 | 0.96 ± 0.01 | 0.97 ± 0.04 |
| | MCI/AD | 0.96 ± 0.02 | 0.97 ± 0.02 | 0.96 ± 0.01 | 0.88 ± 0.06 | 0.92 ± 0.03 |
| | CN/AD | 0.79 ± 0.01 | 0.85 ± 0.01 | 0.86 ± 0.01 | 0.90 ± 0.01 | 0.97 ± 0.03 |

Counterfactual fairness was also assessed across all sensitive attributes for all bias mitigation methods, except for adversarial debiasing. Individual and combined results per subgroup ranged between 0.99 and 1, indicating that this fairness definition was not able to effectively capture existing biases, in contrast to group fairness metrics.

### B. Identification of proxy features based on SHAP analysis

Within the framework of bias mitigation at the preprocessing stage, total brain volume was identified as a proxy feature for all three sensitive attributes (gender, age, race). The SHAP analysis revealed that total brain volume was assigned mean SHAP values of 0.4 and 0.3 for gender and age, respectively, while its corresponding SHAP value for race was two orders of magnitude smaller, indicating its stronger influence on gender and age predictions. Notably, all other ROI volumes in the feature importance ranking by SHAP were regarded as insignificant, compared to total brain volume, for the prediction of sensitive attributes.

## IV. DISCUSSION

The presented fairness analysis identified biases related to age, race, and gender, with the greatest disparities observed in the MCI/AD subproblem. In general, no consistently unprivileged or privileged subgroups emerged across the studied sensitive attributes. With regards to race, in the CN/MCI subproblem, white individuals showed lower TPR and higher FPR, especially when no bias mitigation technique was applied, whereas in MCI/AD, black individuals presented lower TPR and higher FPR values. This can be attributed to the increased imbalance of subgroups in the MCI class, where whites represented 94.3% compared to 87.7% in the overall dataset. Since MCI was considered the positive class in the CN/MCI subproblem, white individuals were more frequently misclassified as positive, whereas in MCI/AD, where MCI represented the negative class, this pattern was reversed. The observed model disparities may be linked to variations in MRI-based predictors of cognition across racial subgroups [25], with atrophy patterns in brain regions associated with distinct disease stages impacting predictions differently across racial groups. For instance, compared to white individuals, African Americans have been shown to exhibit increased hippocampal atrophy and cortical thinning in regions commonly affected by AD [25]. This may explain the higher FPR observed for black individuals in the MCI/AD subproblem, as the model may have overestimated the disease severity in this population.

In terms of gender, no significant biases were identified, except for the MCI/AD subproblem, which exhibited a low equalized odds ratio. It is noteworthy that the low equalized odds ratio in this case was driven by instances of zero values of the FPR ratio, where one gender subgroup had zero FPR, while the FPR of the opposite subgroup was also close to zero. Thus, despite the low equalized odds ratio, no actual disparities existed. ROC was highlighted by the harmonic mean as the best performing bias mitigation technique, since it achieved comparable fairness to other methods while outperforming them in terms of utility metrics. Although women account for two-thirds of AD patients [26], the balanced gender distribution across the dataset and individual classes resulted in fair outcomes in terms of this sensitive attribute. In general, apart from differences in brain structure, women have been shown to exhibit a faster atrophy rate in AD-related brain regions compared to men [27]. In view of these differences, a fairness study of AD disease progression could potentially provide further insights into gender-related biases [12].

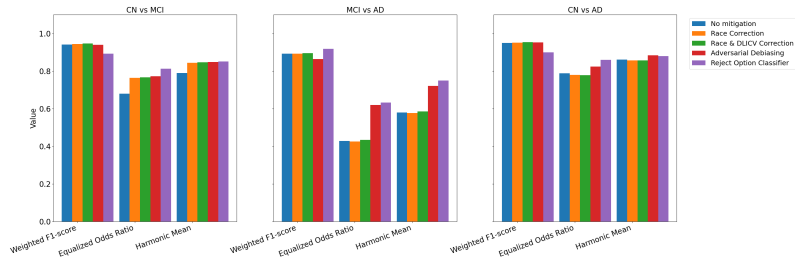

Fig. 2: Results of weighted F1-score, equalized odds ratio, and their harmonic mean across bias mitigation methods for race.

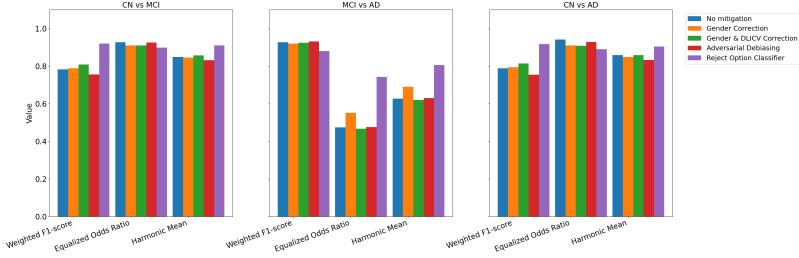

Fig. 3: Results of weighted F1-score, equalized odds ratio, and their harmonic mean across bias mitigation methods for gender.

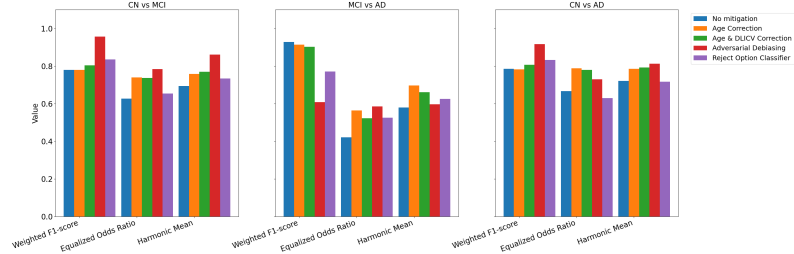

Fig. 4: Results of weighted F1-score, equalized odds ratio, and their harmonic mean across bias mitigation methods for age.

For age, participants over 69 years exhibited higher TPR and FPR, indicating more effective disease detection, but also a tendency for overdiagnosis. This may be related to the fact that individuals under 69 years represented only 27.5% of MCI and 15.9% of AD cases, limiting the model's ability to learn balanced predictions across age groups. This finding also aligned with the observation that normal brain aging can have overlapping patterns of brain atrophy with early stages of AD and, thus, can lead the model to overpredict MCI and AD in older populations.

Among the examined group fairness metrics, balanced accuracy parity and F1-score parity ratios were ineffective in detecting existing biases, as their values were close to 1 across all sensitive attributes and bias mitigation techniques. Even though demographic parity exhibited low values in certain cases, it was not considered a reliable metric able to uncover existing biases, as it disregards the ground truth labels, thus merely reflecting disparities with respect to positive predictions, regardless of their validity. Similarly, counterfactual fairness, by assessing the stability of predictions between data instances and their sensitive feature counterfactuals, without considering the accuracy of predictions, failed to reveal existing biases, which highlighted this fairness definition's limitations in the studied context [28].

On the other hand, equalized odds emerged as a more reliable fairness measure, able to better reflect existing biases in the data, particularly in the context of disease prediction, by accounting both for the TPR and FPR. However, as highlighted by the fairness analysis of gender, it may sometimes indicate the existence of disparities that are not actually present. While calculating equalized odds based on the absolute differences in TPR and FPR rather than the ratios could address this inconsistency, it would hinder the proportional comparisons in terms of this metric across models and bias mitigation methods. Thus, including the TPR, FPR, and FNR calculations of distinct subgroups in the fairness analysis is crucial for a comprehensive understanding of sources of bias.

With respect to the explored bias mitigation strategies, it was observed that linear correction did not significantly increase the equalized odds ratio for race and gender, but it improved overall fairness in terms of age, particularly in the MCI/AD subproblem, where it was highlighted by the harmonic mean as the optimal technique that maximized performance and fairness. However, the incorporation of the total brain volume as a proxy feature in linear correction did not yield significant improvement in fairness or performance across the considered sensitive attributes, suggesting that correction based solely on sensitive attributes was sufficient for removing sensitive attribute-related information. Adversarial debiasing demonstrated satisfactory fairness and performance results across all sensitive attributes and outperformed other methods in many cases with respect to the harmonic mean metric, which indicated its effectiveness in addressing the existing biases. ROC achieved a low harmonic mean in the fairness analysis of age, which was identified as the most biased sensitive attribute, but yielded high harmonic mean values in the case of gender, as it was shown to contribute to improved predictive performance. In cases where bias was present, ROC proved ineffective in mitigating it, as merely adjusting the prediction threshold did not address bias at its source.

## V. CONCLUSIONS

The present study performed a thorough fairness assessment of ML models for MCI and AD diagnosis in a large and diverse dataset, including multiple cohorts, by investigating the effect of existing biases across age, gender, and race. A comprehensive analysis of fairness definitions and metrics was performed and their ability to reliably detect biases was compared, particularly for cases of disease prediction, while various bias mitigation techniques were evaluated. The introduction of the harmonic mean of weighted F1-score and equalized odds ratio yielded consistent results that enabled the reliable comparison of bias mitigation methods with respect to both performance and fairness, by quantifying their trade-off.